\documentclass[lettersize,journal]{IEEEtran}
\usepackage{amsmath,amsfonts}
\usepackage{algorithmic}
\usepackage{algorithm}
\usepackage{array}
\usepackage[caption=false,font=normalsize,labelfont=sf,textfont=sf]{subfig}
\usepackage{textcomp}
\usepackage{stfloats}
\usepackage{url}
\usepackage{verbatim}
\usepackage{graphicx}
\usepackage{blindtext}
\usepackage{cite}
\usepackage[percent]{overpic}
\usepackage{epsfig}
\usepackage{cite}
\usepackage{hyperref}
\usepackage{color,soul}
\usepackage{multirow}
\usepackage{booktabs}
\usepackage{subfig}
\usepackage{mwe,tikz}
\usepackage[percent]{overpic}

\newcommand\copyrighttext{%
	\footnotesize \textcopyright This article was submitted to Soft Robotics Journal on May 2023.
}

\newcommand\copyrightnotice{%
	\begin{tikzpicture}[remember picture,overlay]
		\node[anchor=north,yshift=-10pt] at (current page.north) {\fbox{\parbox{\dimexpr\textwidth-\fboxsep-\fboxrule\relax}{\copyrighttext}}};
	\end{tikzpicture}%
}

\begin{document}
\title{Reconfigurable, Transformable Soft Pneumatic Actuator with Tunable 3D Deformations for Dexterous Soft Robotics Applications}
\author{Dickson Chiu Yu Wong, Mingtan Li, Shijie Kang, Lifan Luo, and Hongyu Yu$^{*}$

\thanks{This work was supported by the Innovation and Technology Commission (project: ITS/036/21FP) of HKSAR, Research Grants Council of Hong Kong under the General Research Fund (Grant No. 16203321), the Foshan HKUST Projects (FSUST21-HKUST03B, FSUST21- FYTRI04B), and was supported in part by the Project of Hetao Shenzhen-Hong Kong Science and Technology Innovation Cooperation Zone (HZQB-KCZYB-2020083).}%

\thanks{Dickson Chiu Yu Wong, Mingtan Li, Shijie Kang and Lifan Luo are with Hong Kong University of Science and Technology (HKUST), Clear Water Bay, Hong Kong. {\tt\small $\{$dcywong, mt.li, skangak, lluoan$\}$@conect.ust.hk}}
\thanks{Hongyu Yu is with Hong Kong University of Science and Technology (HKUST), Clear Water Bay, Hong Kong, and HKUST Shenzhen-Hong Kong Collaborative Innovation Research Institute, Futian, Shenzhen, China {\tt\small hongyuyu@ust.hk}}
\thanks{$^{*}$ Address all correspondence to this author.}
}

\maketitle

\copyrightnotice


\begin{abstract}

Numerous soft actuators based on PneuNet design have already been proposed and extensively employed across various soft robotics applications in recent years. Despite their widespread use, a common limitation of most existing designs is that their action is pre-determined during the fabrication process, thereby restricting the ability to modify or alter their function during operation. To address this shortcoming, in this article the design of a Reconfigurable, Transformable Soft Pneumatic Actuator (RT-SPA) is proposed. The working principle of the RT-SPA is analogous to the conventional PneuNet. The key distinction between the two lies in the ability of the RT-SPA to undergo controlled transformations, allowing for more versatile bending and twisting motions in various directions. Furthermore, the unique reconfigurable design of the RT-SPA enables the selection of actuation units with different sizes to achieve a diverse range of three-dimensional deformations. This versatility enhances the RT-SPA’s potential for adaptation to a multitude of tasks and environments, setting it apart from traditional PneuNet. The paper begins with a detailed description of the design and fabrication of the RT-SPA. Following this, a series of experiments are conducted to evaluate the performance of the RT-SPA. Finally, the abilities of the RT-SPA for locomotion, gripping, and object manipulation are demonstrated to illustrate the versatility of the RT-SPA across different aspects.

\end{abstract}



\begin{figure*}[t]
	\centering
	\includegraphics[width =\textwidth]{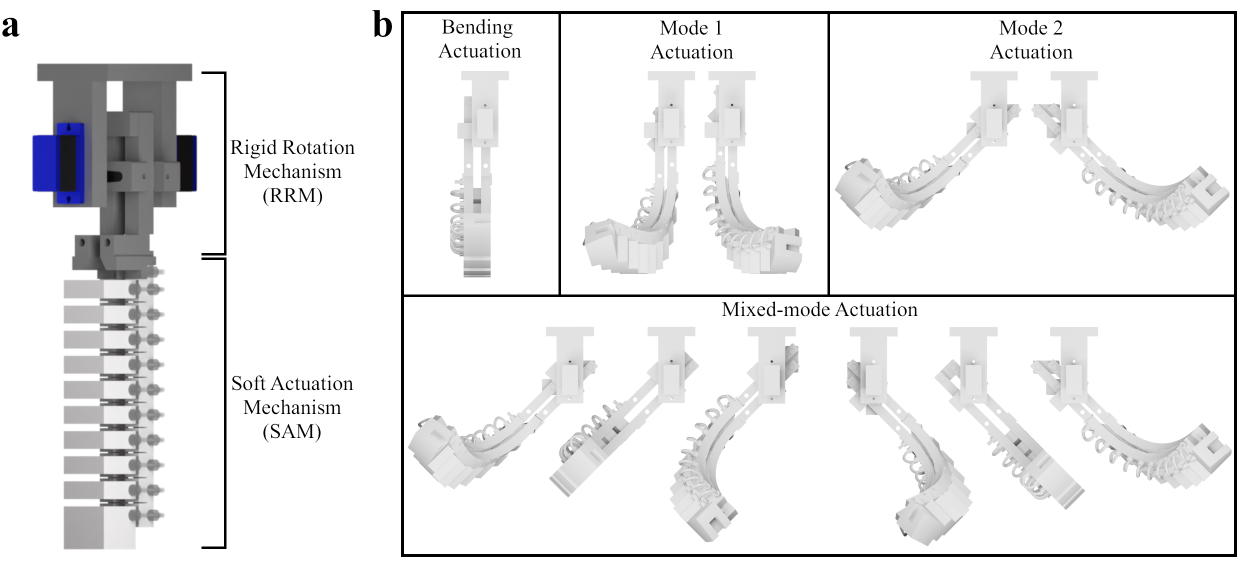}
	\caption{\textcolor{black}{Conceptional design of the RT-SPA. (a) Overall assembly of the RT-SPA, which consists of the Rigid Rotation Mechanism (RRM), tasked with carrying out the structural transformation, and the Soft Actuation Mechanism (SAM), designed to execute actuation upon pressurization following the completion of the transformation. (b) Illustration of the four actuation modes of the RT-SPA. The selection of actuation mode is influenced by the transformation executed by the RRM, which is determined by the operational state of the two embedded servo motors. Subsequent to the transformation, the SAM is pressurized to execute the corresponding bending or twisting actuation.}}
	\label{fig:design_idea}
    \vspace{-1em}
\end{figure*}

\section{Introduction}

Soft Robotics, an emerging subfield in robotics, has attracted interest from diverse research areas in recent years~\cite{rus2015design, REN2021103075}. The intrinsic properties of softness, flexibility, and compliance provide exceptional resolutions to challenges faced by rigid counterparts. Typically, “softness” is defined by material compliance, often described by Young’s moduli~\cite{adem201700016, Chubb_2019}, which leads to the concept of an infinite degree of freedom, enhancing mobility in soft robotic systems. Consequently, soft robotics exhibits advantages in adapting to unknown objects and environments, ensuring safer human/environment-robot interactions, and providing cost-effective fabrication methods, all of which can barely be found in the conventional field of robotics~\cite{aah3690}. Therefore, soft robotics shows promising development across various fields and applications such as gripping~\cite{dma201707035, s21093253}, locomotion~\cite{admt201900837, rsif.2017.0101}, crop-harvesting~\cite{9829727, ZHANG2020105694}, construction~\cite{FIRTH2022104218}, and rehabilitation~\cite{chu2018soft, bardi2022upper}.

Extensive studies have been conducted to explore different actuation methods for soft robotics~\cite{dma201707035, s21093253}, such as soft pneumatic actuator (SPA), tendon-driven actuation, shape-memory alloy (SMA), dielectric elastomer actuator (DEA), and magnetic actuation~\cite{aisy.201900077, admt.202100018, aisy.202100165, admt.202101672, aisy.202000128}. Recently, novel actuation methods have been proposed, such as hydrogel~\cite{LEE2020100258}, liquid crystal elastomer~\cite{aisy.202000148}, liquid metal~\cite{admt.201800549}, and light actuation~\cite{D0MH01406K}. Among all these actuation methods, the SPA is popular in soft robotics because of the characteristics of a large actuation force, fast response time, and simpler structural design. Furthermore, pneumatic systems can be easily implemented, and a safe operation can be guaranteed as no high voltage, high temperature, or strong magnetic field is involved~\cite{adem201700016, Pagoli_2022, 9785890}. In the context of fabrication methods, traditional SPA fabrication methods, such as molding and soft lithography, present limitations in geometric complexity and require multiple steps. With recent advancements in 3D printing, intricate structural designs can be efficiently incorporated, resulting in a streamlined and optimized fabrication process~\cite{STANO2021101079, aisy.202000223}.

Basically, the design of SPAs can be divided into two main categories. The first is the fiber-reinforced soft actuator, whose action mainly focuses on the external braiding pattern and the inextensible fiber~\cite{soro.2015.0001}. As a typical example, the fiber-reinforced bending actuator is capable of generating bending action by the asymmetric expansion formed by the top braiding pattern and the bottom strain-limiting layer~\cite{8752271, soro.2020.0087}. Another example is the well-known Mckibben actuator or pneumatic artificial muscle (PAM). The elongation or contraction motion, determined by the braiding angle, can be generated once the pressure is applied~\cite{1045389X11435435, SOLEYMANI2020620
}. The second category encompasses geometry-determined soft actuators, usually made of elastomeric material. Linear actuators belong to this category, in which the elongation or contraction motion can be generated through symmetric expansion or shrinking of the wavy structure~\cite{8788588, frobt.2019.00034}. Pneumatic Network (PneuNet) represents commonly used actuators in which the bending actuation is generated by the asymmetric expansion of the top chambers~\cite{soro.2016.0030, adfm.201303288, 9244610}. 

In recent years, efforts have been made to advance the dexterity and applicability of SPAs~\cite{scirobotics.abg6049}. Research was conducted on the actuator design to modify the geometry to yield different actuation outcomes. Yeow’s group~\cite{8405577}, Alici’s group~\cite{8452456, soro.2019.0015}, and Gu’s group~\cite{WANG2018131, soro.2020.0039} proposed a twisting actuator design by altering the chamber angle of the PneuNet. Zou’s group proposed an origami-based twisting actuator~\cite{admt.201800429, advs.201901371}. Chen et al.~\cite{9403872} and Xiao et al.~\cite{1.4047989} proposed a twisting actuator based on freeform surface design and helical structure, respectively. Furthermore, Chen et al. designed a bending-twisting actuator using topology optimization~\cite{1.4053159}. However, these modified actions are pre-defined in the fabrication and assembly stage, with no alterations permitted during application.

Researchers have aimed to perform multiple actions by cascading actuators on the system integration level. Fei et al. proposed a stiffness-tunable soft gripper by combining bidirectional soft pneumatic bending actuators~\cite{soro.2018.0015}. Drotman et al. developed a 3D-printed soft actuator by cascading three bellows to perform gripping and locomotive actions in different directions~\cite{8520773}. Wang’s group proposed the combination of linear actuators to carry out different applications through structural design~\cite{9854144, 9632349}. Zheng’s group used soft bending actuators and bistable carbon-fiber-reinforced polymer laminates for rapid shape-morphing~\cite{soro.2020.0200, soro.2019.0195}. Alternatively, some researchers include rigid mechanisms to expand the workspace of SPAs. Zhong et al. proposed a four-fingered gripper that can perform both parallel and concentric gripping by rotating two fingers~\cite{ZHONG2019445}. Huang et al. developed a four-fingered gripper incorporating mechanisms to control the gripper’s opening and individual rotation of all actuators~\cite{huang2020variable}. Cui et al. proposed a four-fingered gripper in which the posture can be controlled by embedded soft elongation and bending actuators~\cite{9380478}. Ye et al. proposed a four-fingered gripper in which a single motor is required to carry out multiple gripping postures~\cite{ye2022design}. In this approach, an increase in overall structural size is unavoidable to increase the degree of freedom of the system.

In this article, a reconfigurable, transformable soft pneumatic actuator (RT-SPA) is proposed. Compared to the conventional SPA, the RT-SPA possesses transformation capabilities, enabling both bending and twisting actions in different directions. Additionally, the RT-SPA’s inherent reconfigurability offers the advantage of utilizing actuation units of various sizes to accomplish tasks with varying degrees of intensity. The rest of the passage is organized as follows. The design idea and the fabrication process of the RT-SPA are discussed. Then, tests and analyses are conducted to assess the functionality of the RT-SPA. Finally, applications of the RT-SPA in locomotion, gripping, and object manipulation are showcased, highlighting its versatility in various contexts. 

\section{Design and Fabrication}

The conceptual design of the RT-SPA is illustrated in Fig.~\ref{fig:design_idea}, consisting of two integral components: the Rigid Rotation Mechanism (RRM) and the Soft Actuation Mechanism (SAM), as shown in Fig.~\ref{fig:design_idea}a. These mechanisms function synergistically to enable the system’s performance. The morphological features of the RT-SPA enable the execution of four distinct actuation modes, as demonstrated in Fig.~\ref{fig:design_idea}b.

\begin{figure*}[t]
	\centering
	\includegraphics[width =\textwidth]{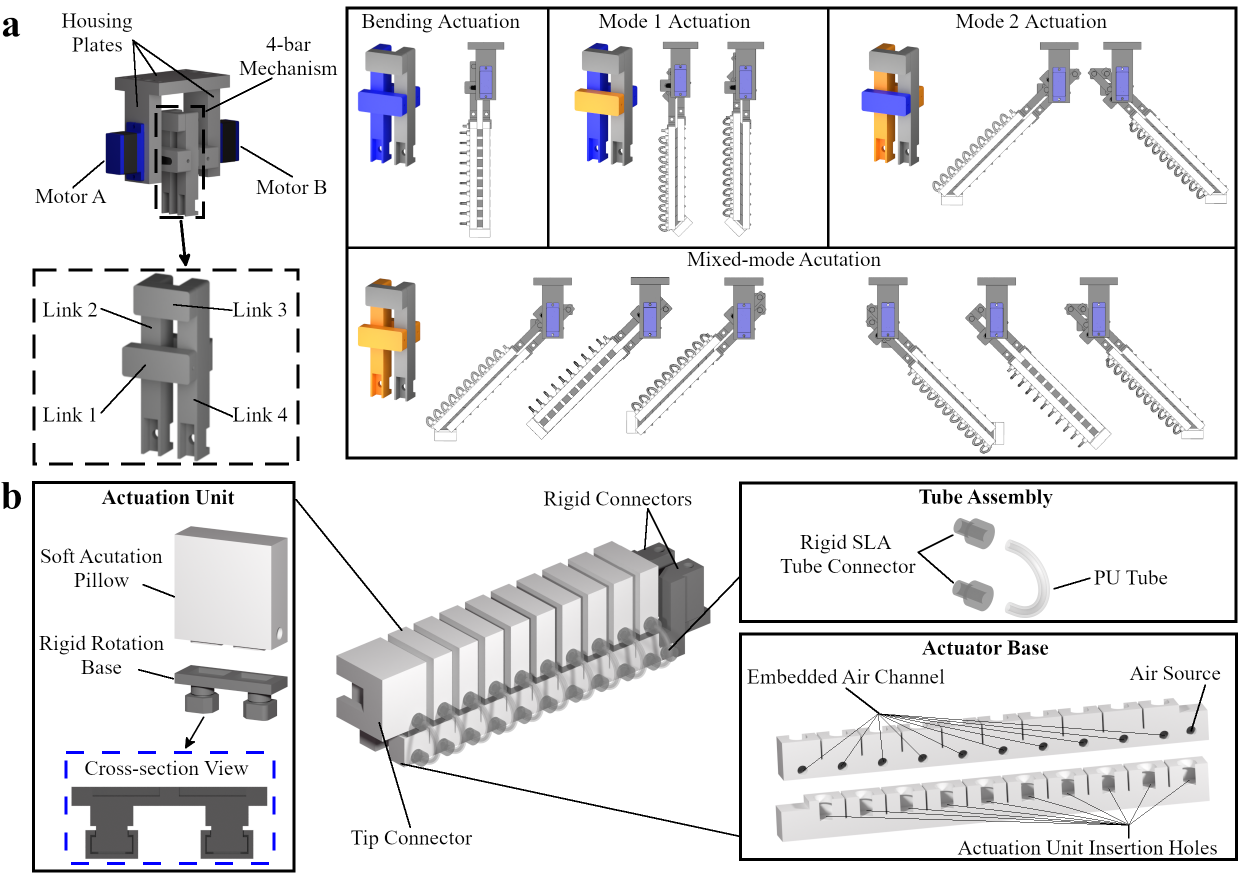}
	\caption{\textcolor{black}{Detailed design of the two RT-SPA mechanisms. (a) Design of the RRM. Demonstrations of RT-SPA, supplemented by the illustration of different actuation modes, are shown on the right side. The 4-bar mechanism in the figure represents its configuration in different modes. A link is designated as ‘fixed’ when the motor is idle and ‘input’ when the motor is actuated. The fixed and input links are colored blue and orange, respectively. (b) Design of the SAM.}}
	\label{fig:fabrciation}
    \vspace{-1em}
\end{figure*}

The fabrication process of the RT-SPA is outlined in Fig.~\ref{fig:fabrciation}, highlighting the use of PLA and Ninjaflex as rigid and soft 3D printing materials, respectively. The RRM employs a conventional parallel 4-bar mechanism, as depicted in Fig.~\ref{fig:fabrciation}a. Two 180-degree servo motors, sharing the same rotational axis and secured by housing plates, are connected to links 1 and 2, capable of rotating $\pm$90 degrees from their initial positions. This arrangement facilitates four distinct rotation modes. In the bending actuation, both motors remain idle, and no transformation takes place, with the RT-SPA functioning as a traditional PneuNet. In Mode 1 actuation, motor A is actuated, making link 1 the input link while motor B holds link 2 fixed, resulting in the rotation of actuation units. In Mode 2 actuation, motor B is actuated with motor A held stationary, designating links 2 and 1 as input and fixed links, respectively. As a result, the Soft Actuation Mechanism (SAM) undergoes bidirectional rotation, preserving a uniform orientation across actuation units. The Mixed-mode actuation can be achieved by executing Mode 1 and Mode 2 actuations simultaneously, thereby enhancing the transformation capabilities of the RT-SPA. 

The design concept of the SAM is depicted in Fig.~\ref{fig:fabrciation}b, consisting of actuation units, SAM bases, tube assemblies, and rigid connectors. Actuation units comprise soft actuation pillows and rigid pillow bases, with the rotation mechanism embedded inside the pillow base. A soft tip connector is integrated at the last actuation unit, enabling the design of end tips with varying shapes for different applications. The soft actuator base is fabricated with the embedded air channel, enabling tube assemblies to connect actuation units to either side of the actuator base for pressurized air supply. Holes on top facilitate actuation unit insertion, and patterned grooves are designed to minimize material resistance during actuation. The tube assembly comprises SLA-printed rigid tube connectors and the PU tube, providing a pathway for pressurized air supply. Finally, rigid connectors are included at the end of actuator bases to secure the SAM assembly to the RRM.

\section{Results and Discussions}

\subsection{Analysis of the Transformation and Actuation Abilities of the SAM}

\begin{figure}[t]
	\centering
	\includegraphics[width=\columnwidth]{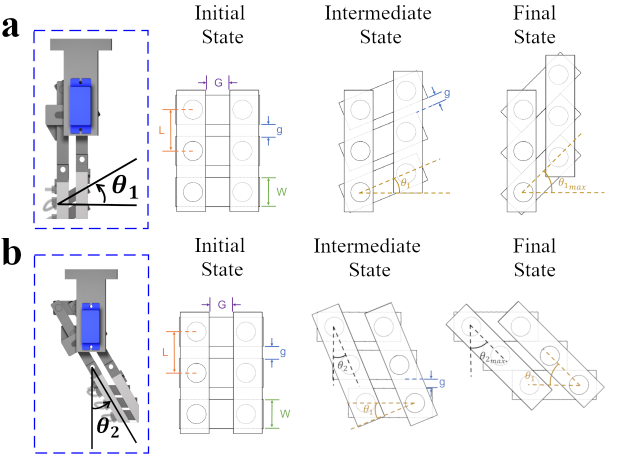}
	\caption{Analysis of the transformation ability of the SAM. The positive notation of the transformation angles $\theta_1$ and $\theta_2$ are shown in the figure. (a) Mode 1 actuation. (b) Mode 2 actuation.}
	\label{fig:transformation_analysis}
	\vspace{-1em}
\end{figure}

In the RT-SPA design, the RRM serves as the source of transformation. Two transformation angles, $\theta_1$ and $\theta_2$, are defined to characterize the actuator’s motion. The angle $\theta_1$ represents the angle between the actuation unit and the line perpendicular to the actuator base, and the angle $\theta_2$ corresponds to the angle between the actuation base and the line parallel to the actuator base. Their notations are defined in Fig.~\ref{fig:transformation_analysis}, where counterclockwise rotation is considered positive. The range of $\theta_1$ and $\theta_2$ is mainly determined by the geometry of the SAM. From the schematics in Fig.~\ref{fig:fabrciation}a, four different rotation modes can be performed according to the actuation status of motors A and B. 

In Mode 1 actuation, the range of $\theta_1$ can be evaluated according to Fig.~\ref{fig:transformation_analysis}. At the initial state ($\theta_1$=0), actuation units are parallel to each other, and a pure bending action can be carried out. Thus, bending actuation can be considered a specialized case for Mode 1 actuation. The gap between actuation units, denoted as $g$, is at maximum and equals to ($L$ – $W$), in which $L$ and $W$ are the center distance between actuation units and the width of the actuation unit, respectively. At the intermediate state (0$<$$\theta_1$$<$$\theta_{1_{max}}$), the transformation of the SAM leads to an increase in $\theta_1$ and a simultaneous decrease in $g$. The value of $\theta_1$ can be related to $W$, $L$, and $g$ by the following geometric equation:

\begin{equation}
    \label{theta_1_intermediate}
    \centering
    \begin{aligned}
        cos\theta_1 = \frac{W+g}{L}
    \end{aligned}
\end{equation}

At the final stage ($\theta_1$ = $\theta_{1_{max}}$), $g$ vanishes when actuation units touch each other. Now the value of the transformation angle attains the maximum value $\theta_{1_{max}}$, which can be calculated by:

\begin{equation}
    \label{theta_1_max}
    \centering
    \begin{aligned}
        cos\theta_1 = \frac{W}{L}
    \end{aligned}
\end{equation}

In the current configuration of the SAM, the values of $\theta_1$ can range from $-\theta_{1_{max}}$ to $\theta_{1_{max}}$. Moreover, the actuator base remains vertical, so the value of $\theta_2$ remains unchanged and is equal to zero. In addition, the distance between actuator bases $G$ must be larger than $g$ to guarantee a proper transformation.

In Mode 2 actuation, the range of $\theta_2$ can be calculated by the same formulations as that of $\theta_1$ in Mode 1 actuation, as shown in Fig.~\ref{fig:transformation_analysis}. Moreover, the bidirectional rotation motion of the SAM, as well as the unchanged orientation of actuation units, lead to a passive change in $\theta_1$, which is equal to $-\theta_2$. 

The transformation ability of the RT-SPA can be further expanded in the Mixed-mode actuation. When the rotation angle of motor B exceeds $\theta_{2_{max}}$, motor A can be used as compensation to increase the motor B angle further, which helps enhance the functionality of the RT-SPA.

\subsection{Actuation Pillow FEM Analysis}

\begin{figure*}[ht]
	\centering
	\includegraphics[width = \textwidth]{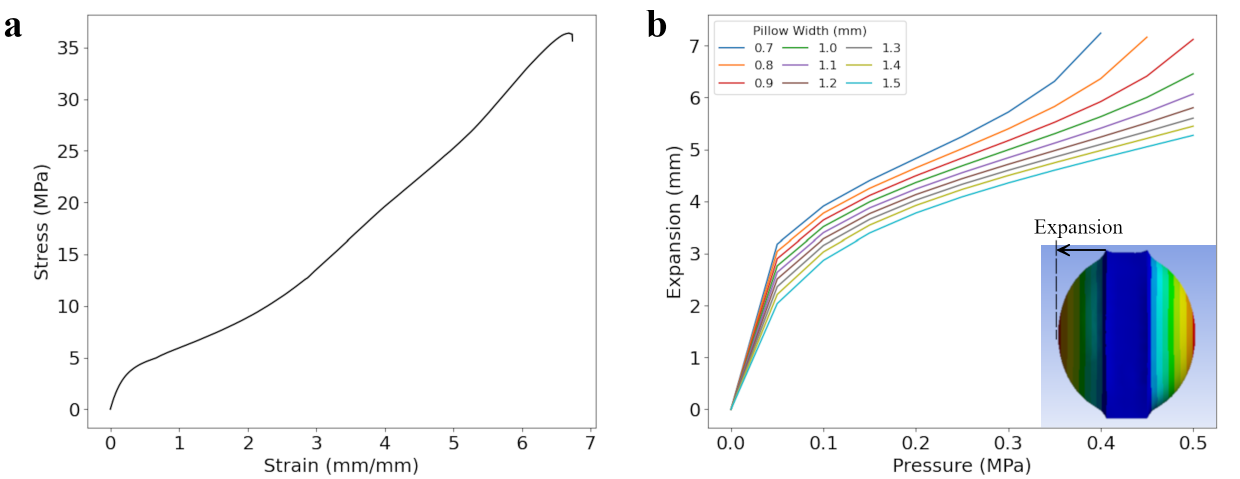}
	\caption{\textcolor{black}{FEM simulation results of the actuation pillow. (a) Stress-strain curve for the Ninjaflex obtained from the uniaxial tensile test. (b) Simulation results of actuation pillows with different front/back wall thicknesses.}}
	\label{fig:FEM}
	\vspace{-1em}
\end{figure*}

The inflatable soft actuation pillows serve as the source of actuation for carrying out the bending and twisting actions and are crucial to the performance of the RT-SPA. Therefore, FEM analysis is carried out to optimize the soft pillow’s design by obtaining the maximum unit expansion, which mainly depends on the thickness of the front and back walls. In our current design, the area of the front and back walls is set to 20x20mm, and the thickness of the side walls is set to 1.2mm to accommodate the overall RT-SPA size. In addition, the pillow is designed to withstand a maximum pressure of 0.5MPa.

A uniaxial tension test was first carried out to characterize the Ninjaflex material according to the ASTM D412 Type C standard. The resultant stress-strain curve is shown in Fig.~\ref{fig:FEM}a. The curve was then imported to ANSYS, and the Ogden 1\textsuperscript{st} order model was chosen as the hyperelastic model for the simulation. The simulation results in Fig.~\ref{fig:FEM}b show that the 0.9mm wall thickness gives the optimal expansion under the maximum design pressure of 0.5MPa. Nevertheless, to secure the 3D-printing quality and the usage of the pillow, 1mm is finally selected as the pillow’s front and back wall thickness.

\subsection{Workspace Analysis}

\begin{figure*}[ht!]
	\centering
	\includegraphics[width = 0.9\textwidth]{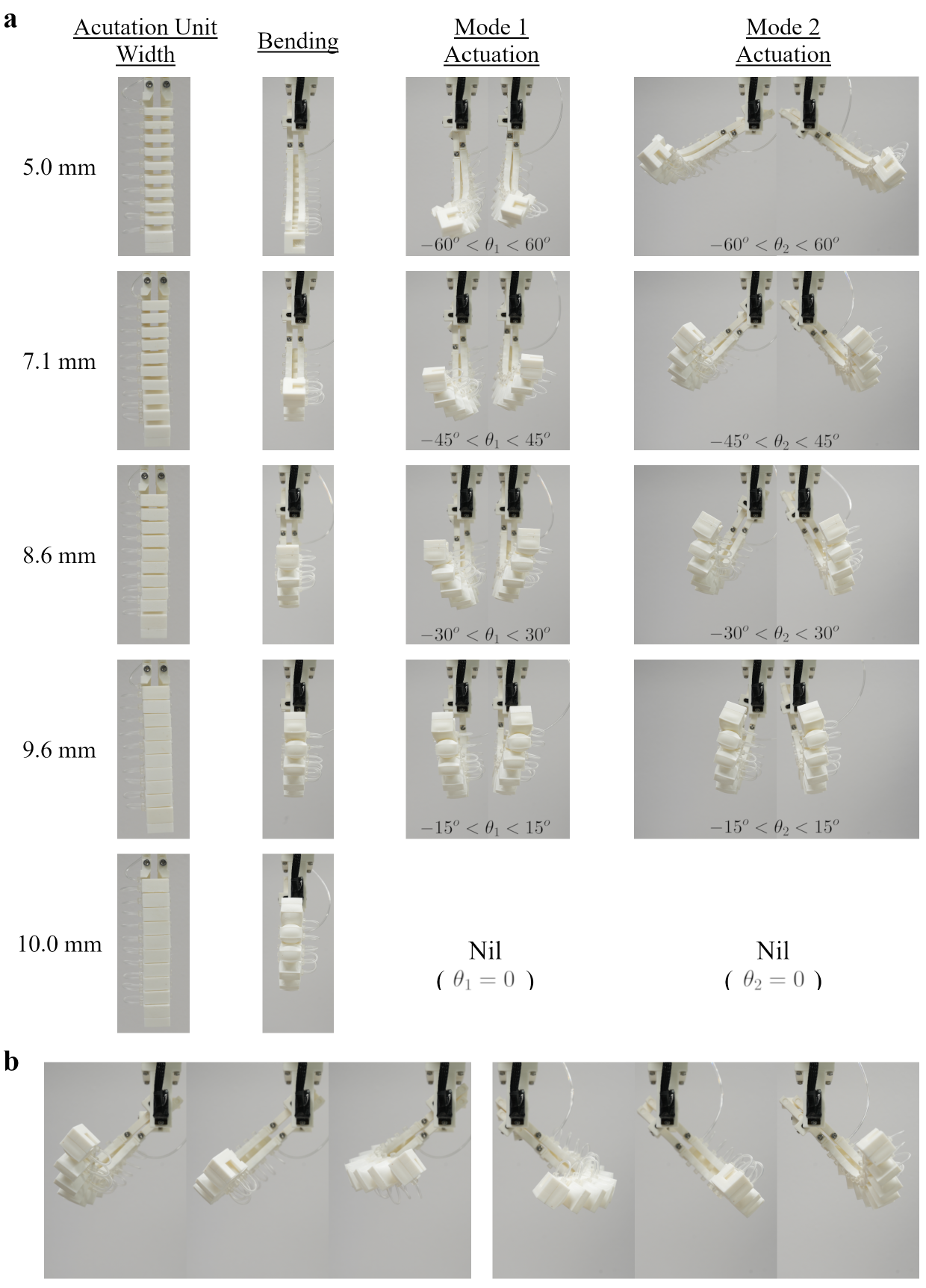}
	\caption{\textcolor{black}{RT-SPA used in the analyses. The actuation unit width and the maximum transformation angle are shown in the figure. (a) Demonstration of actuation performance under pure bending, Mode 1 and Mode 2 actuation. (b) Demonstration of actuation performance under Mixed-mode actuation by the 7.1mm actuation unit.}}
	\label{fig:actuator_demo}
\end{figure*}

\begin{figure*}[ht!]
	\centering
	\includegraphics[width = 0.85\textwidth]{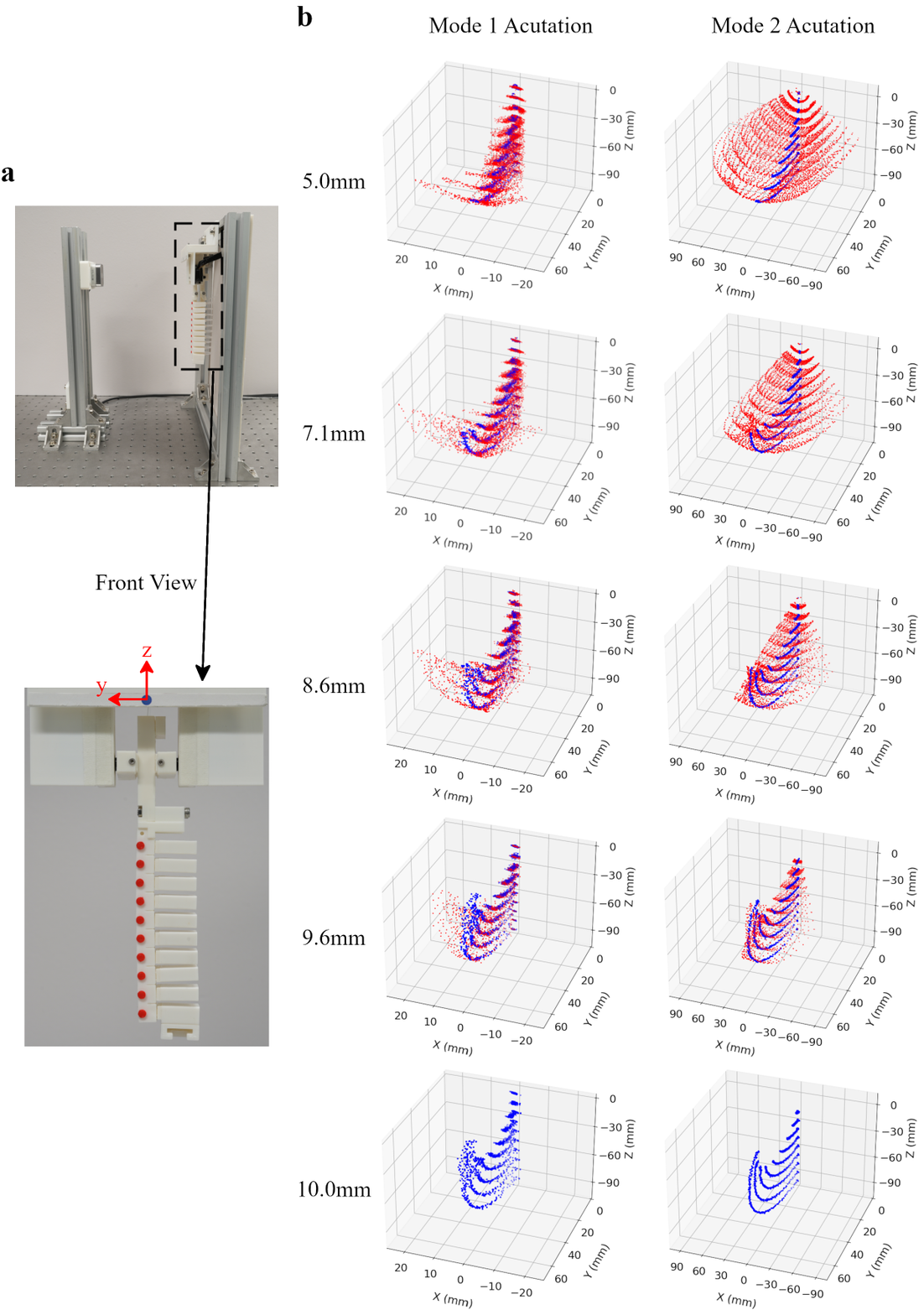}
	\caption{\textcolor{black}{Workspace analysis of the RT-SPA. (a) Experimental setup of the 3D imaging measurement system. The blue point was set as the origin. The coordinate system is marked in the figure, with the x-axis directed to the page. (b) Results of workspace measurement of the 5 tested RT-SPA in Mode 1 and Mode 2 actuations. The red dots represent the workspace generated by the RT-SPA. The blue dots represent the workspace generated by the RT-SPA in the bending Mode, which can be considered the equivalent conventional PneuNet actuator with the same design parameters.}}
	\label{fig:workspace_test}
\end{figure*}

Several analyses were carried out to evaluate the performance of the RT-SPA. Five different RT-SPAs with different actuation unit widths were examined. The center distance between actuation units across all samples was kept constant in these samples. The actuation performances of these samples under bending, Mode 1, and Mode 2 actuations can be seen in Fig.~\ref{fig:actuator_demo}a. In addition, the demonstration of the Mixed-mode actuation with the 7.1 mm-width soft actuation unit, is shown in Fig.~\ref{fig:actuator_demo}b. 

One of the most important characteristics of the RT-SPA is the integration of servo motors in the RRM to control the shape morphing of the actuator and carry out the desired bending and twisting motions. Compared to the conventional bending SPAs, the workspace is significantly increased. Here, a 3D imaging measurement system was set up to measure the resultant workspace generated by the RT-SPA in Mode 1 and Mode 2 actuations, as shown in Fig.~\ref{fig:workspace_test}a. The RT-SPA was held down on the stand. Dots with two different colors were used in the measurement. A blue dot was marked on the RRM housing plate and served as the origin of the measurement. Ten red dots were evenly marked on the SAM and were used as trackers to trace the position changes of the SAM during actuation. An RGB-D camera (Intel RealSense D405) was used for image acquisition. For Mode 1 actuation, motor A was set to rotate from 0 to $\theta_{1_{max}}$ at a 5-degree interval. At each $\theta_1$, the RT-SPA was pressurized from 0 to 0.5MPa continuously, and images were taken by the RGB-D camera simultaneously. A Python program was written to calculate the coordinates of red dots in 3D space. The same process was also carried out to measure the workspace for Mode 2 actuation. As the motion is symmetric, data is mirrored about the y-z plane to generate the complete point cloud. The resultant point clouds in x-y and x-z planes for Mode 1 and Mode 2 actuations of the five tested RT-SPA are shown in Fig.~\ref{fig:workspace_test}b. The detailed workspace representations in the x-y and the y-z planes are provided in the supplementary information.

The workspace of the conventional PneuNet and the RT-SPA are denoted by blue and red data points, respectively. A careful examination of the figure reveals that the workspace of the RT-SPA is greatly enhanced in comparison to the single bending action of the conventional PneuNet actuator, which is attributed to the transformation capabilities of the RT-SPA. In addition, several observations can be made from the above result. First, the results above show that Mode 1 actuation primarily serves to tune the bending and twisting actuation, and Mode 2 actuation is mainly used for changing the direction of the actuation. Moreover, when the actuators are in the bending Mode ($\theta_1$=0), the maximum attainable bending angle increases with $W$. As $L$ is kept constant throughout the experiment, increasing $W$ leads to decreases in $g$. Thus, decreasing amounts of expansion are required to overcome $g$ before the contact between actuation units occurs, which increases the bending performance of the actuator. Furthermore, considering the Mode 1 actuation, an increase in $W$ results in a greater allowance of rotation for actuation units, leading to an increase in twisting performance. These findings suggest a compensatory relationship between the attainable bending and twisting actions on an actuation unit with varying $W$ under constant $L$ conditions.

\subsection{Force Analysis}

\begin{figure*}[ht!]
	\centering
	\includegraphics[width = \textwidth]{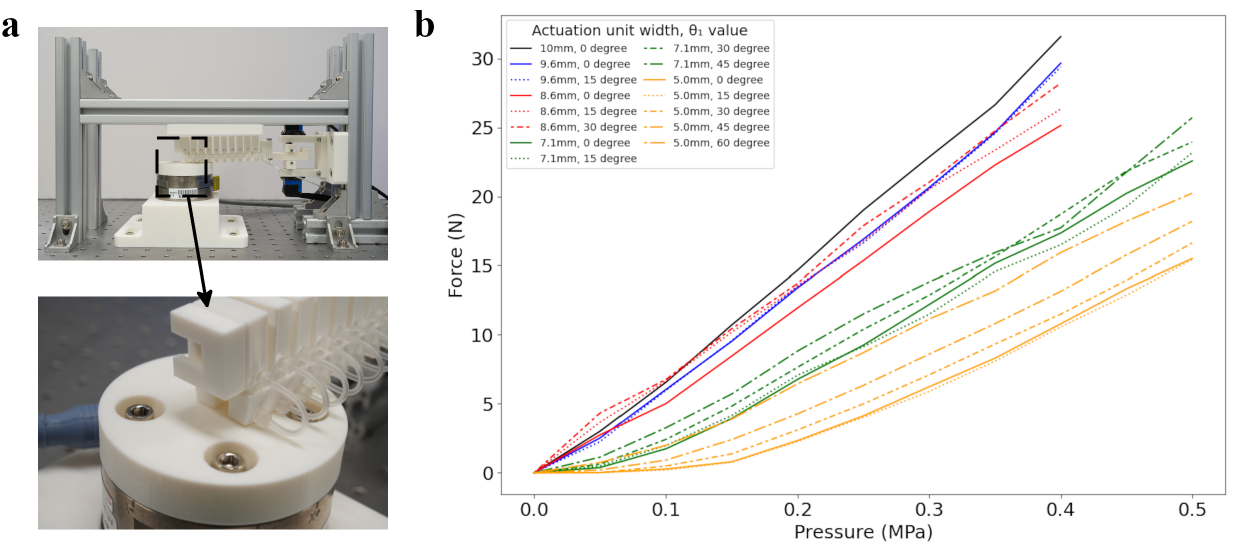}
	\caption{\textcolor{black}{Blocking force test. (a) Test setup. The RT-SPA position is tuned to ensure that the tip is located at the center of the force sensor. (b) Test results.}}
	\label{fig:force_test}
	\vspace{-1em}
\end{figure*}

\begin{figure*}[ht!]
	\centering
	\includegraphics[width = \textwidth]{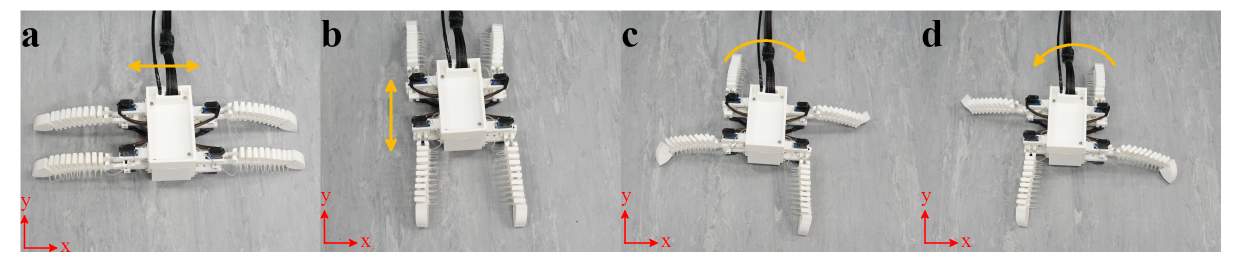}
	\caption{\textcolor{black}{Demonstration of the RT-SPA ability in locomotion. The coordinates are marked on the figures, with the z-axis pointing out of the page. (a) Linear motion along the x-axis. (b) Linear motion along the y-axis. (c) Clockwise rotation above the z-axis. (d) Counter-clockwise rotation above the z-axis.}}
	\label{fig:locmomtion}
	\vspace{-1em}
\end{figure*}

As one of the most important metrics to evaluate the performance of the soft actuator, the blocking force of the RT-SPA was tested by the setup shown in Fig.~\ref{fig:force_test}a. The force sensor (ATI 9105-TIF-GAMMA) was used to measure the blocking force. The RT-SPA was fixed on one end, and its position was tuned to ensure that the actuator tip was located at the center of the force sensor. As depicted in the workspace analysis, the tuning of bending and twisting actions is mainly influenced by the Mode 1 actuation. Therefore, Mode 1 actuation was carried out to measure the forces in this experiment. The transformation angle $\theta_1$ was increased from 0 to $\theta_{1_{max}}$ with a 15-degree increment. At each $\theta_1$, the RT-SPA was actuated from 0 to the maximum withstanding pressure without damaging the structure. The results are shown in Fig.~\ref{fig:force_test}b.

Two observations can be made from the results. First, in the case of bending, the maximum force increases with the actuation unit width. As described in the workspace analysis, an increase in $W$ leads to a decrease in the expansion of the actuation unit required to overcome $g$, which in turn directly enhances the blocking force and the overall bending performance. Second, under the same actuation unit width, a gradual increment in the blocking force is observed as $\theta_1$ increases. While actuation units rotate, $g$ decreases; however, the concurrent increase of the offset between centers of actuation units weakens the effect of the expansion, and the resulting blocking force.

In summary, the relationship between actuation unit width and transformation angle highlights the interplay between these variables in determining the actuator’s bending and twisting capabilities, as well as the resulting blocking force. Understanding these relationships is essential for optimizing the design of the RT-SPA and its application in various soft robotics tasks.

\section{Applications}

In this section, three different applications are demonstrated to showcase the universality of the RT-SPA. The 7.1mm-width actuation unit was chosen for the following demonstrations.

\subsection{Locomotion}

A four-legged robot was fabricated to showcase the locomotion ability of the RT-SPA, as shown in Figure 8. The robot could move in different directions depending on the orientation of $\theta_1$ and $\theta_2$  on the four RT-SPAs. In Fig.~\ref{fig:locmomtion}a, $\theta_1$ and $\theta_2$ of all actuators are set to zero, and the robot is placed such that the actuators are parallel to the x-axis. In this configuration, the robot can execute motion along the x-axis. By implementing the Mixed-mode actuation to re-orient all actuators in parallel to the y-axis, the robot can move linearly in the positive and negative y-axis, as shown in Fig.~\ref{fig:locmomtion}b. On the other hand, the robot can also perform rotational motions, as shown in Fig.~\ref{fig:locmomtion}c and Fig.~\ref{fig:locmomtion}d. The motion is achieved by setting two opposing actuators perpendicular to each other through the Mixed-mode actuation, while executing Mode 1 actuation on the remaining two opposing actuators. Therefore, positive and negative rotations about the z-axis were performed accordingly.

\subsection{Gripping}

\begin{figure*}[ht!]
	\centering
	\includegraphics[width = \textwidth]{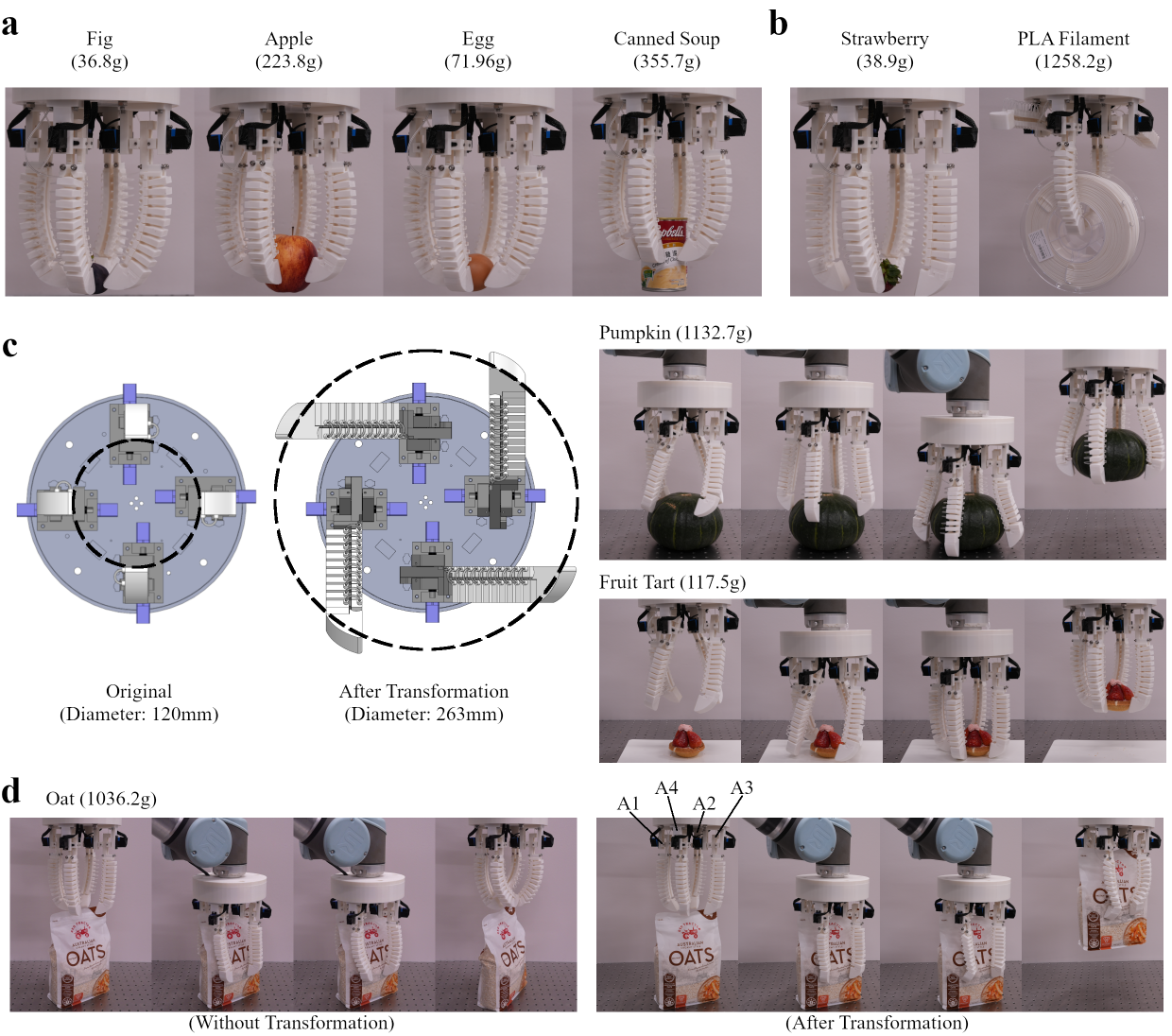}
	\caption{\textcolor{black}{Demonstration of the RT-SPA ability in gripping actions. The tested objects and their respective weights are labeled in the diagram. (a) Conventional four-fingered cage gripping. (b) Two-fingered gripping. Depending on the size of the object, this type of gripping can be carried out with or without transformation. (c) Enhanced-opening four-fingered concentric gripping to grasp objects with a size larger than the gripper opening. The left diagram illustrates the comparison of the gripping opening before and after transformation. (d) Pseudo-parallel four-fingered gripping.}}
	\label{fig:gripping}
	\vspace{-1em}
\end{figure*}

A four-fingered concentric gripper with a 120mm opening has been fabricated to demonstrate the ability of the RT-SPA as a versatile and adaptive gripper, as shown in Fig.~\ref{fig:gripping}. The gripper showcases its effectiveness in grasping a variety of objects under the conventional cage gripping configuration, as illustrated in Fig.~\ref{fig:gripping}a.

Three additional gripping modes are discovered based on the uniqueness of the transformation ability of the RT-SPA. The first gripping mode is the two-fingered parallel gripper, as depicted in Fig.~\ref{fig:gripping}b. For the small object, such as the strawberry shown in the figure, no transformation is required for the gripper, and only two opposing RT-SPA are actuated to realize the two-fingered gripping. When dealing with the object, such as the PLA filament spool shown in the figure, the four-fingered concentric gripper can be transformed into a two-fingered gripper by retracting the two opposing actuators by the Mixed-mode actuation and carrying out the hook grip.

The second gripping mode is the four-fingered enhanced-opening concentric gripper, as illustrated in Fig.~\ref{fig:gripping}c. While conventional SPAs may be capable of withstanding large loading, the opening of the gripper may not be sufficiently large to conform to objects, like the pumpkin shown in the figure. The RT-SPA can be transformed to enlarge the gripper opening by carrying out the Mixed-mode actuation on all four actuators, as demonstrated in the left diagram of Fig.~\ref{fig:gripping}c, and the gripper opening diameter increases from 120mm to 263mm. Thus, proper object conformation and grasping can be achieved with a size greater than the gripper opening. Additionally, the enhanced-opening method is beneficial when handling delicate objects, such as the fruit tart shown in the figure. By enlarging the gripper opening, the RT-SPA ensures proper positioning and prevents any damage to the fragile object before carrying out the grasping process.

The third gripping mode is the four-fingered pseudo-parallel gripper, as shown in Fig.~\ref{fig:gripping}d. Initially, the concentric gripper was unable to grasp flat objects, such as the bag of oats in the diagram. After the Mode 1 actuation was carried out on all actuators, actuators A1 and A2 acted as if they were parallel to actuators A3 and A4 when actuated. Thus, the pseudo-parallel gripper was able to grasp the bag of oats successfully.

\subsection{Object Manipulation}

\begin{figure*}[ht!]
	\centering
	\includegraphics[width = \textwidth]{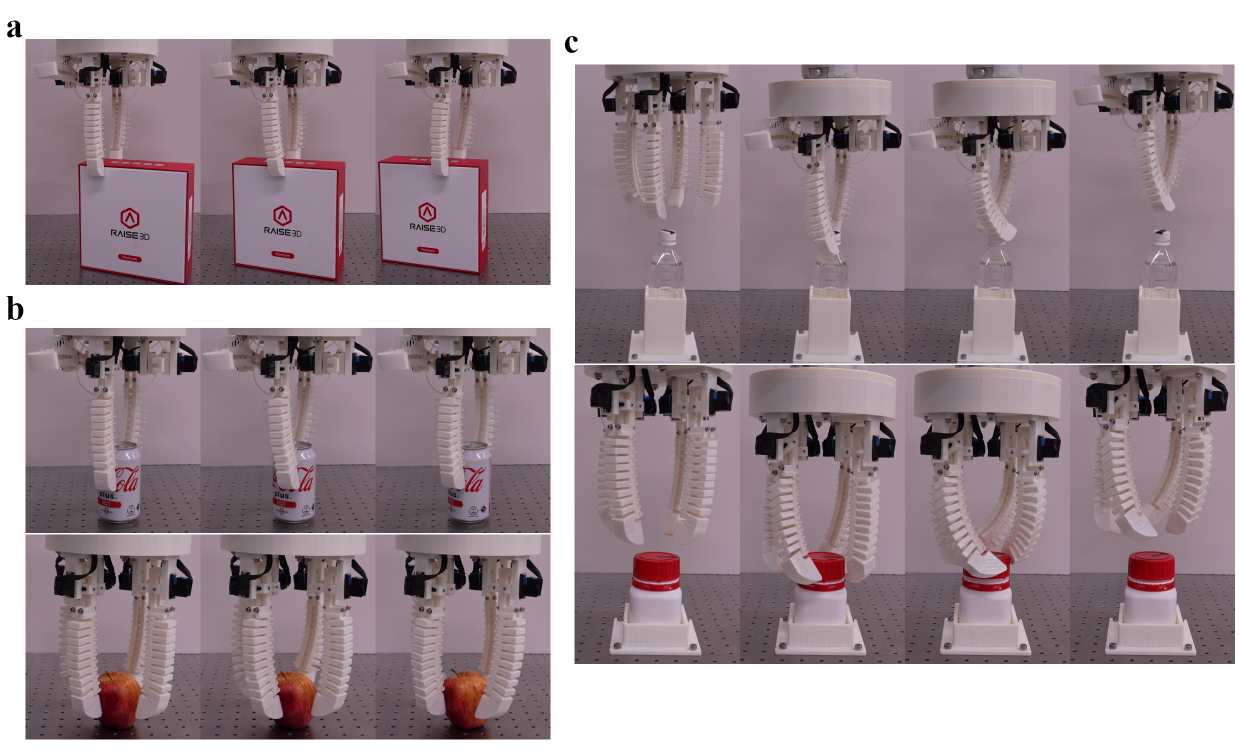}
	\caption{\textcolor{black}{Demonstration of the RT-SPA ability in object manipulation. (a) Objection translation by a two-fingered gripper. (b) Object rotation by parallel movement with two actuators (top) and four actuators (bottom) requires no vertical motion. (c) Object rotation by twisting with two actuators (top) and four actuators (bottom). The gripper is required to move up and down to achieve the action.}}
	\label{fig:object_manipulation}
	\vspace{-1em}
\end{figure*}

The same four-fingered concentric gripper can be further used to demonstrate the manipulation abilities of the RT-SPA, as shown in Fig.~\ref{fig:object_manipulation}. In this context, three different manipulation techniques are demonstrated. The first is the linear object translation by the two-fingered gripper, as shown in Fig.~\ref{fig:object_manipulation}a. Two opposing actuators are first retracted by the Mixed-mode actuation. Then, the remaining actuators are moved and actuated in the Mixed-mode actuation to move the object in a translational manner. 

The second example is the object rotation by parallel actuator movement, which can be carried out by either a two-fingered or four-fingered gripper, as depicted in Fig.~\ref{fig:object_manipulation}b. Actuators are moved and actuated in Mixed-mode actuation, and no up-and-down movement of the gripper is required to perform the manipulation.

Finally, the object rotation by twisting is performed, which can also be carried out by either two-fingered or four-fingered grippers, as shown in Fig.~\ref{fig:object_manipulation}c. Actuators are moved and actuated under Mode 1 actuation with a designated value of $\theta_1$, combined with the up and down motion of the gripper to carry out the manipulation.

\section{Conclusion}

In this article, a novel soft pneumatic actuator RT-SPA is proposed with both reconfigurability and transformability, enabling a controlled 3D deformation to different extents. We designed and tested various RT-SPAs with different actuation unit widths. The results demonstrate that our proposed actuator can carry out bending and twisting motions in a controlled manner. Several analyses were conducted to evaluate the performance of the RT-SPA. Finally, applications of RT-SPA on gripping, locomotion, and object manipulations are demonstrated to showcase its versatility.

In the future, research will focus on the development of the analytical model for the RT-SPA, which will enable a more in-depth understanding of its underlying principles and facilitate further improvement of its performance. Additionally, we will explore sensor integration to augment the actuator’s maneuverability, thereby broadening its scope of applicability even further.

\section*{ACKNOWLEDGEMENT}
The authors would like to thank HKUST Shenzhen-Hong Kong Collaborative Innovation Research Institute, Futian, Shenzhen for supporting this project. Also, the authors would like to acknowledge Mark ELLWOOD for proofreading this review and helping check the spelling and grammar.


\bibliographystyle{ieeetr}
\bibliography{references.bib}


\newpage

\end{document}